\documentclass[letterpaper, 10 pt, journal, twoside]{ieeetran}
\IEEEoverridecommandlockouts

\usepackage{graphics} 
\usepackage{epsfig} 
\usepackage{subfigure}

\usepackage{mathptmx} 
\usepackage{times} 
\usepackage{amsmath} 
\usepackage{amssymb} 
\usepackage[ruled,linesnumbered]{algorithm2e}
\usepackage{multirow}

\usepackage{cite}
\usepackage[colorlinks=true,linkcolor=black,urlcolor=black,citecolor=black]{hyperref}
\usepackage{bm} 
\usepackage{algorithmic}

\usepackage{color}
\usepackage{hyperref}

\title{Offline-Online Learning of Deformation Model for Cable Manipulation with Graph Neural Networks}

\author{Changhao Wang$^{1}$, Yuyou Zhang$^{2}$, Xiang Zhang$^{1}$, Zheng Wu$^{1}$, Xinghao Zhu$^{1}$\\ Shiyu Jin$^{1}$, Te Tang$^{3}$, and Masayoshi Tomizuka$^{1}$
\thanks{$^1$University of California, Berkeley, CA, USA.
{\tt\footnotesize changhaowang, xiang\_zhang\_98, zheng\_wu, zhuxh, jsy, tomizuka@berkeley.edu}}
\thanks{$^2$Shanghai Jiao Tong University, Shanghai, China.
{\tt\footnotesize oct.10@sjtu.edu.cn}}
\thanks{$^3$FANUC Advance Research Laboratory, Union City, CA, USA.
{\tt\footnotesize Te.Tang@fanucamerica.com}
}
}

\begin{document}

\maketitle

\begin{abstract}
Manipulating deformable linear objects by robots has a wide range of applications, e.g.,  manufacturing and medical surgery. 
To complete such tasks, an accurate dynamics model for predicting the deformation is critical for robust control. In this work, we deal with this challenge by proposing a hybrid offline-online method to learn the dynamics of cables in a robust and data-efficient manner. In the offline phase, we adopt Graph Neural Network (GNN) to learn the deformation dynamics purely from the simulation data. Then a linear residual model is learned in real-time to bridge the sim-to-real gap.
The learned model is then utilized as the dynamics constraint of a trust region based Model Predictive Controller (MPC) to calculate the optimal robot movements. The online learning and MPC run in a closed-loop manner to robustly accomplish the task. Finally, comparative results with existing methods are provided to quantitatively show the effectiveness and robustness.
Experiment videos are available at
\href{https://msc.berkeley.edu/research/deformable-GNN.html}{https://msc.berkeley.edu/research/deformable-GNN.html}.
\end{abstract}

\begin{IEEEkeywords}
Manipulation Planning, Dual Arm Manipulation, Deformable Object Manipulation
\end{IEEEkeywords}

\section{Introduction}
Manipulating deformable linear objects, especially cables, has a wide range of applications. For example, in factory manufacturing, both stator winding and cable harnessing require cables to be assembled in a precise manner. In medical surgery, steering needles or catheters are essential for vascular pathology treatment and minimally invasive surgery. 
In order to complete the above tasks automatically, robustly manipulating cables to the desired shape remains a fundamental problem for robotics. However, due to the high dimensionality and nonlinearity, it is challenging to obtain an accurate dynamics model for precise planning and control.

The finite element method (FEM) serves as a powerful tool to model cables. Through discretizing the objects into small elements, the deformation can be derived by solving a set of partial differential equations. FEM can accurately represent the dynamics of deformable objects with fine tessellation~\cite{brenner2008mathematical}. However, it is usually computationally expensive to solve, thus posing a challenge to use it for real-time robotic manipulation tasks.

Aside from the FEM models, data-driven approaches are often utilized to learn the non-linear deformation dynamics~\cite{yu2021adaptive,hu20193,yan2020learning}.
Recently, many works have attempted to use deep neural networks (DNN) to learn such dynamics. However, the characteristic of deformable objects is complex, and it is found challenging for those general-purpose network structures to capture the cable behaviors.

Recent studies on graph neural networks (GNN) bring a new structure for model learning. By viewing particles of the object as the graph vertices, the dynamics is modeled as the interaction between the vertices pairs~\cite{sanchez2020learning}. The future object state can be predicted through a sequence of `message passing' blocks, which mimic the transmission of the interaction from one graph vertex to others. Compared with those general-purpose network structures, the GNN encodes the prior knowledge on how the interaction may transmit inside the deformable objects.
However, as most of the machine learning based approach does, the GNN model relies on simulation for data generation. Thus, the sim-to-real gap still poses a significant challenge for the model to be adopted in robotic manipulation tasks.

\begin{figure}
\centerline{\includegraphics[scale = 0.3]{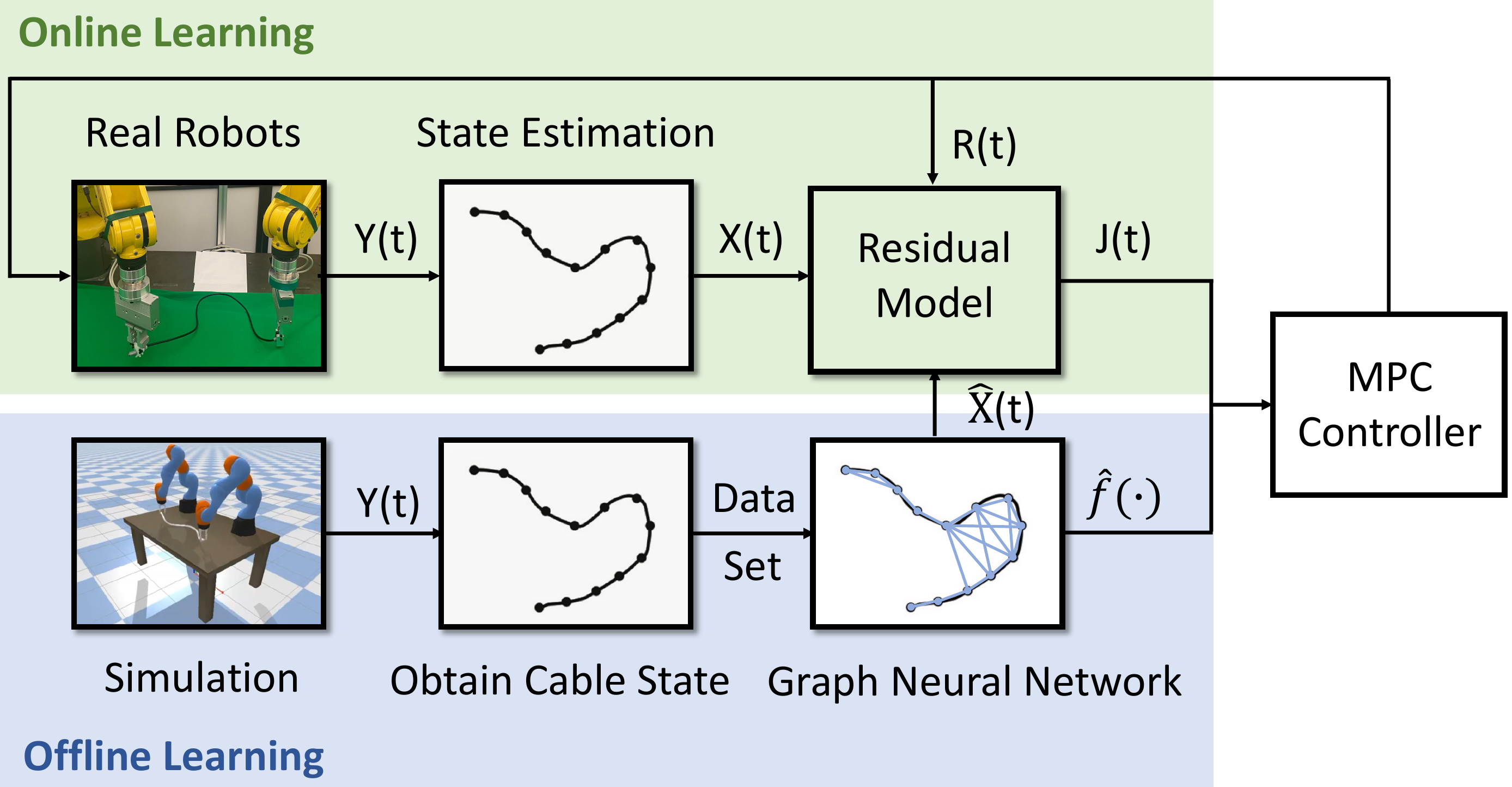}}
\caption{The proposed cable manipulation framework combines offline model learning with online residual learning. In the offline phase, we trained a GNN with simulated data as the global model of the cable. In the online phase, a local residual model that estimates the GNN prediction error is learned in real-time to reduce the sim-to-real gap. The learned models are then sent to an MPC controller to obtain the optimal robot motion.
}
\label{fig_framework}
\end{figure}

To tackle the above issues, we propose a dynamics learning framework that consists of an offline GNN model and an online residual model, as shown in Fig.~\ref{fig_framework}. In the offline phase, we learn rough global graph dynamics from the simulation data. In the online phase, we further refine the local predictions through a time-varying linear residual model that estimates the error of the GNN output. In contrast with existing sim-to-real approaches, our method does not require pre-collecting real-world data. The linear residual model can be learned simultaneously as the robot executes. With this framework, we can efficiently capture the global deformation behaviors without sacrificing local accuracy. 
For manipulation, a trust region based MPC formulation with the learned model is proposed to optimize the robot movement. 
In summary, the main contributions are listed as follows:
\begin{itemize}
    \item Combine the GNN with an online linear residual model for robust model learning.
    \item Formulate a trust region based model predictive controller to optimize the robot movements.
    \item Demonstrate the effectiveness of the proposed method through comparative simulations and experiments.
\end{itemize}

\section{Related Works}
\subsection{Global Deformation Model Learning}

Learning-based approaches approximate the cable dynamics from the collected data set. Nair et al.~\cite{nair2017combining} proposed a predictive model learned from step-by-step images of a rope collected during human demonstration.
Yan et al.~\cite{yan2020self} proposed to utilize bi-directional LSTM to model chain-like objects. By recurrently applying the same LSTM block along the cable, we can enforce how the interaction will propagate inside the model.
Similarly, the graph neural network arises as a novel structure to learn the deformation dynamics. By explicitly modeling the vertices and edges, authors of~\cite{li2018learning} demonstrated the effectiveness of the proposed dynamic particle interaction networks (DPI-Nets) on fluids and deformable foam manipulation tasks.
The graph network structure proposed in~\cite{sanchez2020learning} further improved the generalizability in terms of prediction steps and particle scalability. Moreover, Pfaff et al.~\cite{pfaff2020learning} generalized the graph structure to mesh-based dynamics, which proved to be effective for more complex simulation tasks.
\subsection{Sim-to-real Transfer for Model Learning}

While the neural networks can capture complex behaviors of deformable objects, those methods typically require sim-to-real methods as complements in case of parameter changes or uncertainties encountered in the real-world experiment.
Matching the simulator parameters with real-world parameters and directly applying the learned model to the real world is the most common way to utilize the model. However, the learned model may suffer from uncertainties in the real world.
Another approach is to randomize several aspects of the domain to provide enough variability in training~\cite{tobin2017domain,chebotar2019closing,ramos2019bayessim}. As a result, the learned model can cover a large fraction of scenarios. 
Fine-tuning the learned model with real-world data is also demonstrated to be effective in closing the sim-to-real gap~\cite{yan2020self}. By collecting new data in the real world and re-training the network, we can adapt the model to the new scenario robustly. 

Combining the offline model with a residual model is also beneficial to resolve the sim-to-real gap~\cite{ajay2018augmenting,kloss2017combining}. Since the offline model does not need to match reality accurately, training in simulation is more efficient and more manageable.

\subsection{Online Local Model Estimation}
Instead of learning the global model and transferring it to the real world, online approximating a local model is another powerful way for cable manipulation.
Visual servoing is a commonly used method for online cable manipulation. Visual servo approaches approximate the local linear deformation model by iteratively updating the deformation Jacobian~\cite{navarro2013visually,zhu2018dual,zhu2021vision}, which represents the mapping between the robot end-effector's velocity and the object state. Compared with offline global model learning approaches, there are no sim-to-real gaps, and it can generalize to different objects and scenarios.
Jin et al.~\cite{jin2019robust} further improved the robustness of the online linear model under sensor noises and occlusions. However, as the linear model's expressiveness is limited, it is challenging for the online methods to manage large deformations.

\subsection{Offline-Online Learning for Robust Manipulation}

Though the online model estimation methods are useful in certain tasks, it is still desirable to take advantage of the more powerful offline global model.
As demonstrated in~\cite{yu2021adaptive}, the
offline learning provides the initial guess of the linear Jacobian matrix, and in the online phase, the Jacobian matrix is further updated with an adaptive control law. Distinct from the sim-to-real methods introduced in the previous subsection, this combination does not require additional online data collection. The sim-to-real gap can be instantaneously resolved as the robot executes.
While the motivation of our proposed framework is similar to the motivation of~\cite{yu2021adaptive}, our algorithm is unique in several aspects: 1) we utilize GNN to capture the global model instead of providing the initial guess of the local Jacobian, 2) we online learn a residual model for refinement, 3) we construct an optimization-based MPC controller to obtain optimal robot motions within a long horizon.

\section{Proposed Framework}

In this section, we introduce the proposed framework for cable manipulation as shown in Fig.~\ref{fig_framework}. We mainly focus on the task of precisely shaping a cable to the desired curvature in a 2D and uncluttered environment.
A robust state estimation algorithm for cables is introduced first. Then the cable dynamics are approximated by a combination of an offline graph neural network (GNN) and an online residual model. In the end, a model predictive controller (MPC) is utilized to control the robot to manipulate the cable to desired states.

\subsection{Cable State Estimation}
\label{subsection:state_estimation}
Estimating the state is essential for dynamics learning and manipulation. 
As shown in Fig.~\ref{fig_framework}, a cable is represented by a series of key points $X(t)=[x_1(t),x_2(t),\cdots,x_N(t)]^T$, where $x_i(t)\in \mathbb{R}^{1\times 2}$ denotes the 2D position of the i-th key point at time step $t$. The objective of state estimation is to estimate the position of each key point at each time step from the dense, noisy, and occluded point clouds $Y(t)=[y_1(t),y_2(t),\cdots,y_S(t)]^T$ perceived by stereo cameras.

Following our previous work on structure preserved registration (SPR)~\cite{tang2018track}, we regard the perceived point clouds $Y(t)$ as samples from a Gaussian Mixture Model (GMM), whose centroids are the key point positions $X(t)$. According to the Bayes' theorem, the probability of point $y_s^t$ to be sampled from the mixture model can be expressed by:
\begin{align}
p(y_s^t) = \sum_{n=1}^{N+1} p(n)p(y_s^t | n) 
\end{align}
where $p(n)$ denotes the weight of the $n$-th mixture component, and $p(y_s^t|n)$ is the corresponding probability that $y_s^t$ is sampled from the mixture component. As shown below, we assume all Gaussians share the same weight, and we add a uniform distribution to account for noise and outliers.
\begin{equation}
p(n) = \left\{
\begin{array}{cc}
	(1-\mu)\frac{1}{N},  & n = 1,\dots, N  \\
	\mu, & n = N+1  \\
\end{array}
\right.
\end{equation}
\begin{equation}
p(y_s^t | n) = \left\{
\begin{array}{cc}
	\mathcal{N} (y_s^t ; x_n^t, \sigma^2 \mathbf{I}) ,  & n = 1,\dots, N  \\
	\frac{1}{S}, & n = N+1  \\
\end{array}
\right.
\end{equation}

The problem is then converted to a Maximum Likelihood Estimation (MLE) problem, where we want to optimize the mixture centroids in order to maximize the log likelihood $L$ that the point cloud $Y(t)$ is sampled as shown in (\ref{eqn_log}).
\begin{align}
L(x_n^t, \sigma^2 | Y^t) & = log \prod_{s=1}^{S} p(y_s^t)\\
& = \sum_{s=1}^{S} log(\sum_{n=1}^{N+1}p(n)p(y_s^t|n))
\label{eqn_log}
\end{align}

The minimization can be achieved via performing E-M algorithm on the log likelihood function. Since state estimation is not the main contribution of this paper, we omit the derivation and refer readers to~\cite{tang2018track} and~\cite{tang2018framework} for details.

\subsection{Offline Model Learning with Graph Neural Networks}
\begin{figure} [!ht]
\centerline{\includegraphics[scale = 0.26]{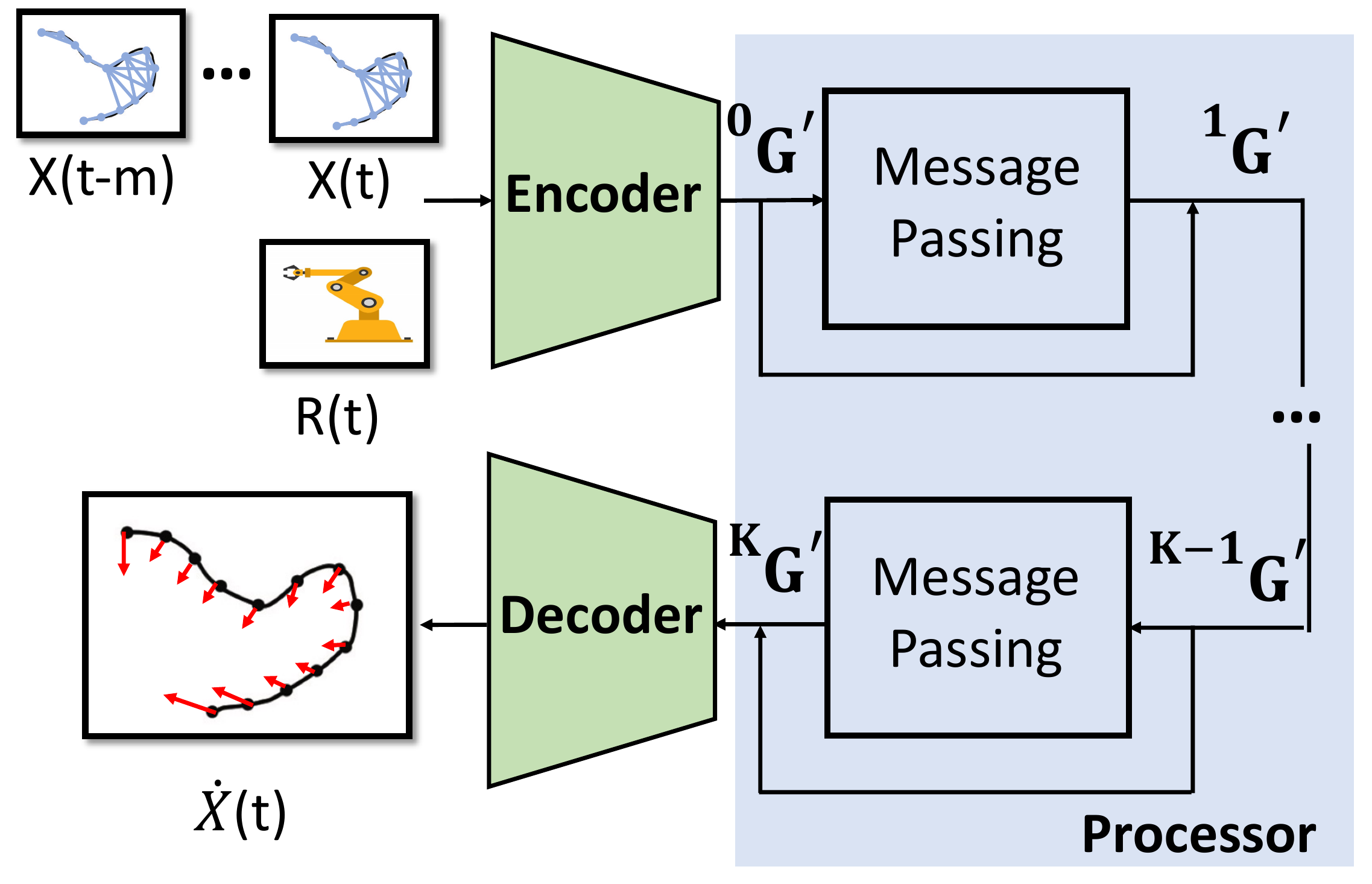}}
\caption{The structure of the Graph Neural Network (GNN). The offline GNN takes in a graph that represents the state of the cable and outputs the prediction of the cable movements.}
\label{fig_graph_dynamics}
\end{figure}
Graph neural networks (GNN) is demonstrated to be effective to represent complex dynamics~\cite{sanchez2020learning}. In this paper, we adopt a similar idea from~\cite{sanchez2020learning} to learn the graph dynamics of the cable. 

We assume that the connection between the robot end-effector and the grasping point is fixed throughout task executions.
The dynamics $f(\cdot)$ is then modeled by the interactions among the cable key points and the robot end-effectors as shown in (\ref{eqn:dynamics}). The dynamics takes in $m+1$ previous cable states from $X(t-m)$ to $X(t)$and the current robot end-effector velocities (translations in $x$, $y$ and rotation around $z$) $ R(t)=[r_1(t),r_2(t),\cdots,r_Q(t)]^T$, where $r_i(t) \in \mathbb{R}^{1\times 3}$ is the end-effector velocities of the i-th robot, and $Q$ is the number of robots. 
\begin{equation}
    \dot{X}(t) = f(X(t-m:t), R(t))
    \label{eqn:dynamics}
\end{equation}

As studied in~\cite{sanchez2020learning}, the dynamics can be captured via a graph $G=(V,E)$, where the graph vertices $V=[v_1,v_2,\cdots,v_N]$ correspond to the key points, and the graph edges $E$ correspond to the interactions between the key points pair. As shown in Fig.~\ref{fig_graph_dynamics}, the features of the $i$-th graph vertex $v_i$ consist of a sequence of previous key point positions $[x_i(t), x_i(t-1), \cdots,x_i(t-m)]$, and the robot control input at that point. 
For the graph edge, it models the relative movement between vertices $e_{i,j} = \|x_i(t)-x_j(t)\|$. An edge is constructed if two vertices are within a `connective radius'.

To learn such a graph, we follow the GNS network structure~\cite{sanchez2020learning}, which contains three steps -- encoding, processing, and decoding:

\begin{figure} [!ht]
\centerline{\includegraphics[scale = 0.25]{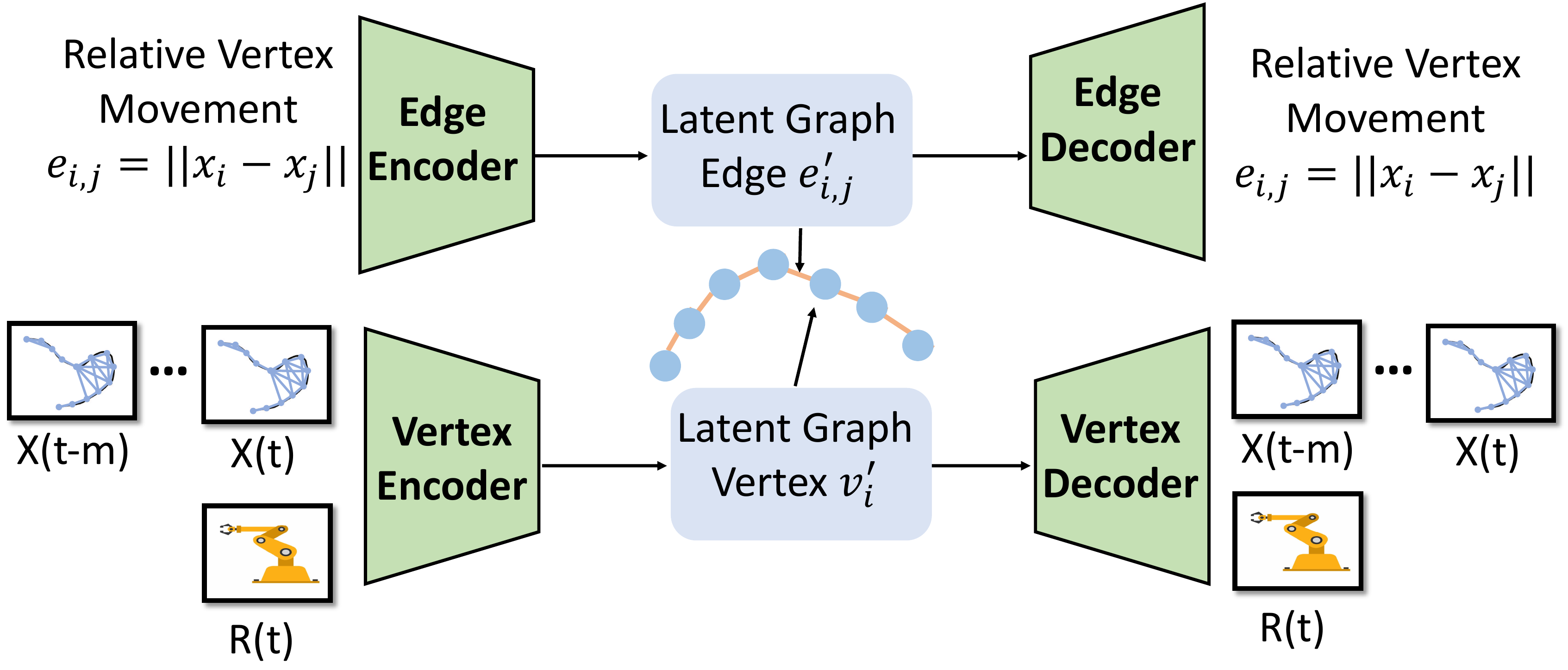}}
\caption{Illustration of the graph encoder and decoder. By minimizing the reconstruction loss in (\ref{eqn:reconsturction}), the encoder can project the high dimensional graph representation to a latent space.
}
\label{fig_auto_encoder}
\end{figure}

\textbf{Encoding:} Instead of operating in the high dimensional space, we project the graph $G$ to a low dimensional representation $G^\prime=(V^\prime,E^\prime)$ for efficiency. Graph vertices auto-encoder and graph edges auto-encoder are trained, respectively as shown in Fig.~\ref{fig_auto_encoder}. The reconstruction error in~(\ref{eqn:reconsturction}) is minimized to make the network learn the optimal latent space that can capture all the original information. $\phi^v$, $\phi^e$ denote the encoder for the graph vertices and edges, and $\psi^v$, $\psi^e$ are the corresponding decoders.
\begin{equation}
\begin{aligned}
    {\phi^v}^*, {\psi^v}^* & = arg \underset{\phi^v,\psi^v}{\text{min}} \|v_i - (\phi^v \circ \psi^v)(v_i)\|^2 \\
    {\phi^e}^*, {\psi^e}^* & = arg \underset{\phi^e,\psi^e}{\text{min}} \|e_{i,j} - (\phi^e \circ \psi^e)(e_{i,j})\|^2
\end{aligned}
\label{eqn:reconsturction}
\end{equation}

\begin{figure} [!ht]
\centerline{\includegraphics[scale = 0.36]{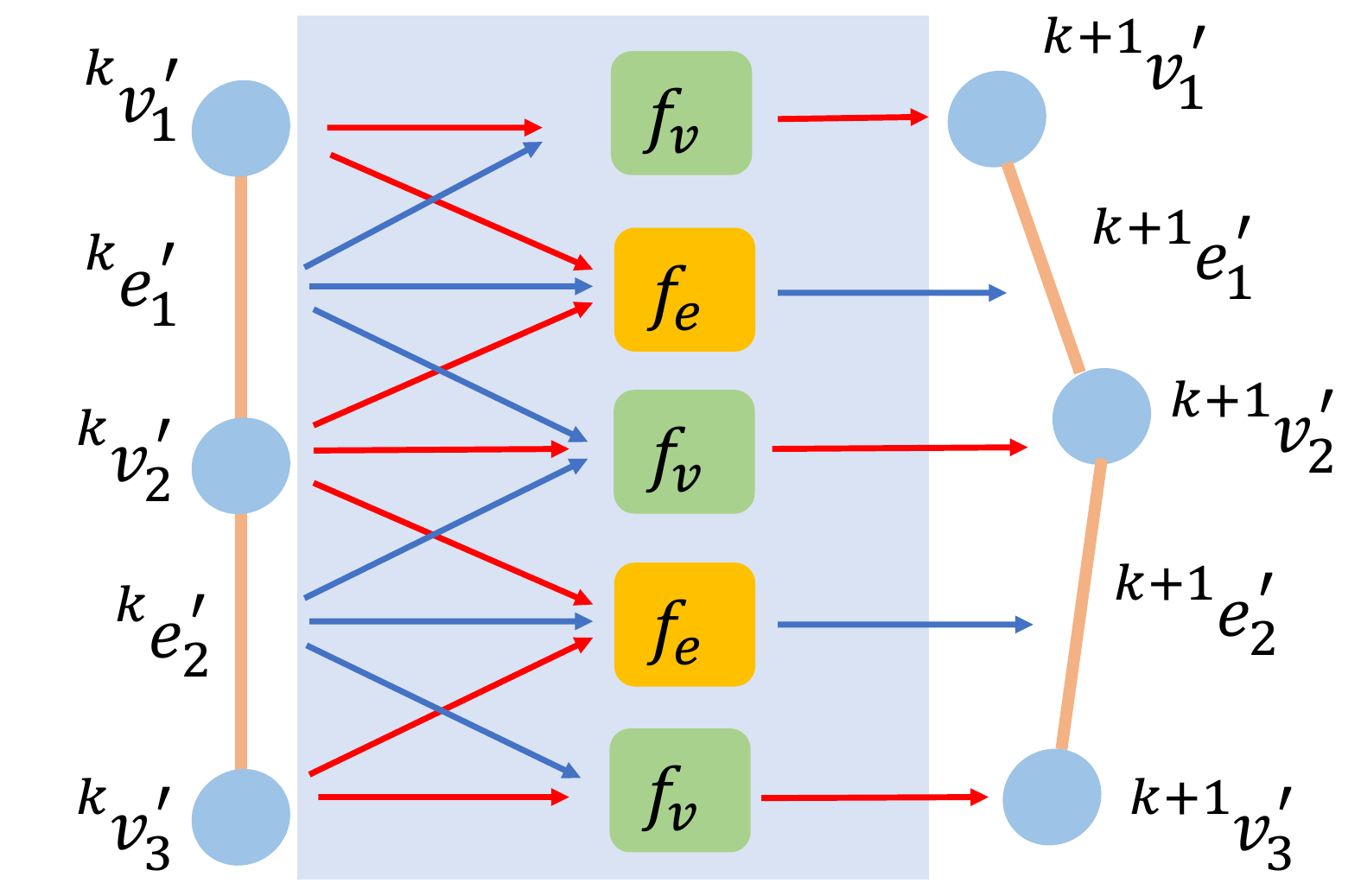}}
\caption{Illustration of the message passing block. Each message passing block transmit the interaction from one graph vertex to others through the graph edge.  
}
\label{fig_message_passing}
\end{figure}

\textbf{Processing:} 
The processor mimics the dynamics of the cable by computing the interaction between vertices in the latent graph.
The message passing block in Fig.~\ref{fig_message_passing} is to propagate the graph vertex, and edge features based on the current graph state and the robot control input as shown in (\ref{eqn:message_passing}).
\begin{equation}
\begin{aligned}
^{k+1}v_i^\prime & = f^v(^{k}v_i^\prime, \sum_j {^{k}e_{i,j}^\prime}) \\
^{k+1}e_{i,j}^\prime & = f^e(^{k}e_{i,j}^\prime, {^{k}v_i^\prime}, {^{k}v_j^\prime})
\end{aligned}
\label{eqn:message_passing}
\end{equation}
where $f^v(\cdot)$ and $f^e(\cdot)$ are the graph vertex network and graph edge network, and $^{k+1}v_i^\prime$, $^{k+1}e_{i,j}^\prime$ denote the latent vertex/edge features after the $k$-th message passing block. 

\textbf{Decoding: } As the inversion of the encoder, the decoder converts the latent graph representation back to get the prediction of the cable movement $\dot{X}(t)$. The weight is directly taken from the decoder part of the auto-encoder $\psi^v$ and $\psi^e$.

\subsection{Online Residual Model Learning}
The offline GNN model is able to learn the global nonlinear dynamics and provide a rough prediction. 
However, due to the sim-to-real gap, the training data from simulation may not accurately capture the dynamics of the physical object.
To overcome the sim-to-real gap and the generalization issues, we propose to online learn a residual model to correct the local predictions.
As shown in (\ref{eqn:residual_model}), $\delta \dot{X}(t)$ is the error between the actual cable state increment $\dot{X}(t)$ and the predicted cable state increment $\widehat{\dot{X}}(t)$ given by the offline learned dynamics denoted as $\widehat{f}(\cdot)$.  For simplicity, we use $X$ to represent the history key point trajectory $X(t-m:t)$. Given the history cable trajectory $X$, we assume that the ground truth residual model is linear with respect to $R(t)$ inside $\|R(t)\|\leq \epsilon$, where $J(t)$ denotes the local residual dynamics. For simplicity, we omit the bias term. It can be considered without much modification.
\begin{equation}
\begin{aligned}
    \delta \dot{X}(t) & = \dot{X}(t) - \widehat{\dot{X}}(t) \\
    & = f(X, R(t)) - \widehat{f}(X, R(t)) \\
    & = R(t) J(t), \quad  \|R(t)\|<\epsilon
\label{eqn:residual_model}
\end{aligned}
\end{equation}

To estimate the residual, we online collect $\delta \dot{\textbf{X}}$ and the corresponding robot end-effector velocities $\textbf{R}$ during the task execution as shown in (\ref{eqn:online_dataset}).
\begin{equation}
\begin{aligned}
    \delta \dot{\textbf{X}} & = \begin{bmatrix}
\delta \dot{X}(t-m) &
\cdots & 
\delta \dot{X}(t-1) &
\delta \dot{X}(t) 
\end{bmatrix}^T \in \mathbb{R}^{(m+1) \times 2N} \\
\textbf{R} & = \begin{bmatrix}
 R(t-m) & 
\cdots &
 R(t-1) &
 R(t)
\end{bmatrix}^T \in \mathbb{R}^{(m+1) \times 3Q}
\end{aligned}
\label{eqn:online_dataset}
\end{equation}

Based on the collected data $\delta \dot{\textbf{X}}$ and $\textbf{R}$, we solve a regularized least squares (ridge regression). The problem can be decomposed to $3Q$ independent ordinary ridge regression problem and solved in parallel. $\delta \dot{\textbf{X}}_n$ denotes the $n$-th column of $\delta \dot{\textbf{X}}$, and similarly, $J_n(t)$ represents the $n$-th column of $J(t)$.
\begin{equation}
\begin{aligned}
\widehat{J}(t) & = arg \underset{J(t)}{\text{min}}
& & \|\delta \dot{\textbf{X}}- \textbf{R} J(t)\|_F^2 + \lambda \|J(t)\|_F^2\\
& = \sum_{n=1}^{6N} arg \underset{J_n(t)}{\text{min}}
& & \|\delta \dot{\textbf{X}_n}-  \textbf{R} J_n(t) \|_2^2 + \lambda \|J_n(t)\|_2^2
\end{aligned}
\end{equation}

The solution is given below, where $\lambda$ is the regularization weight, and $I$ denotes the identity matrix.
\begin{equation}
    \widehat{J}_n^*(t) = (\delta \dot{\textbf{X}_n}^T \delta \dot{{\textbf{X}_n}} + \lambda I)^{-1}\delta \dot{\textbf{X}_n}^T  \textbf{R}
\label{eqn:least_square_solution}
\end{equation}

\subsection{Model Predictive Control for Manipulation}
To utilize the learned model for manipulation, we adopt the idea of model predictive control (MPC).

\begin{equation}
\begin{aligned}
\min_{ R(0:h)} \quad & \sum_{t=1}^{h+1} \|X(t)  - X_d \|_2^2\\
\textrm{s.t.} \quad & X(t+1) = X(t) + \dot{X}(t) \Delta t\\
&\dot{X}(t) =  \widehat{f}(X, R(t)) +  R(t)\widehat{J}(t)\\
& \| R(t)\| \leq \epsilon\\
& t = 0, 1, \cdots, h
\label{eqn_MPC_formulation}
\end{aligned}
\end{equation}

As shown in (\ref{eqn_MPC_formulation}), the optimization variables are the robot's end-effector velocities within a horizon $h$, and the objective function is to minimize the shape position error. The constraints include the learned dynamics, where $\Delta t$ denotes the sample time of the dynamics. More importantly, we consider movement limit constraints to ensure that the learned model (offline and online) is in effect in the optimized region. Borrowing the idea from the trust-region optimization, we update the trust-region size according to the ratio $\rho$ between the actual shape error reduction $\Delta e_{actual}$ and the predicted shape error reduction $\Delta e_{pred}$ as shown in (\ref{eqn_trust_region_ratio}). The pseudo-code is summarized in Alg.~\ref{alg_MPC}, where the trust-region size will expand/shrink if $\rho$ is greater/smaller than a threshold.
\begin{equation}
    \rho = \frac{\Delta e_{actual}}{\Delta e_{pred}}
\label{eqn_trust_region_ratio}
\end{equation}

\begin{algorithm}
\caption{Trust Region Based Model Predictive Control}
\begin{algorithmic}[1] 
\label{alg_MPC}
\REQUIRE Initialize $R$, $\epsilon$
\WHILE{$\|X(t) - X_d\| \geq e$}
    \STATE $R \leftarrow$ Solve the optimization in (\ref{eqn_MPC_formulation})
    \STATE $X(t+1) \leftarrow$ Execute the robot and obtain the new cable state
    \STATE $\rho \leftarrow$ Calculate the ratio according to (\ref{eqn_trust_region_ratio})
    \IF {$\rho \geq \eta^{+}$}
        \STATE $\epsilon \leftarrow \tau^+ \epsilon$ 
    \ELSIF {$\rho \leq \eta^-$}
        \STATE $\epsilon \leftarrow \tau^- \epsilon$ 
    \ENDIF
    \ENDWHILE

\end{algorithmic}
\end{algorithm}

During the task execution, the MPC controller iteratively solves for the optimal robot movements and sends the results to the robot for execution. Similar as our previous work~\cite{wang2021trajectory}, the optimization is solved by IPOPT~\cite{wachter2006implementation}, which implements a primal-dual interior-point linear search algorithm. The local convergence is guaranteed as proved in~\cite{wachter2005line}. We empirically demonstrated the effectiveness through simulation and experiments.

\section{Simulation and Experiments}
\label{section:simulation}

\begin{table}
\centering
\caption{Parameter Values of the Proposed Approach}
        \label{table2}
\begin{tabular}{  |l| l| } 
                \hline
                Simulator Parameters & Value \\ \hline
                \qquad Cable Length & $1m$ \\
                \qquad Cable Diameter & $0.01m$ \\
                \qquad Spring Elastic Stiffness & $4\times 10^3$ \\ 
                \qquad Spring Damping Stiffness & $2\times 10^3$ \\
                \qquad Spring Bending Stiffness & $3\times 10^3$ \\ 
                \qquad Global Scale & $2$ \\ \hline
                GNN Model Parameters & Value\\ \hline
                Encoder & \\
                \qquad Number of Hidden Layers & $2$ \\
                \qquad Size of Hidden Layers & $128$ \\
                \qquad Activation Function & ReLU \\
                Processor & \\
                \qquad Time Window Size ($m$) & $5$ \\
                \qquad Number of Message Passing ($k$) & $10$ \\ 
                \qquad Connective Radius & $0.2m$ \\
                \qquad Number of Hidden Layers & $2$ \\
                \qquad Size of Hidden Layers & $128$ \\
                \qquad Activation Function & ReLU \\
                Decoder & \\
                \qquad Number of Hidden Layers & $2$ \\
                \qquad Size of Hidden Layers & $128$ \\
                \qquad Activation Function & ReLU \\ \hline
                GNN Training Parameters & Value \\ \hline
                \qquad Batch Size & $1$ \\
                \qquad Learning Rate & $10^{-4}$ to $10^{-6}$ \\ \hline
                MPC Parameters & Value\\ \hline
                \qquad MPC Horizon ($h$) & $5$ \\
                \qquad Dynamics Sample Time ($\Delta t$) & $1s$ \\
                \qquad Regularization Weight ($\lambda$) & $10$ \\
                \qquad Initial Trust Region Size & $0.05m$ \\
                \qquad $\eta^{+}$, $\eta^{-}$ & $0.8$, $0.4$ \\
                \qquad $\tau^{+}$, $\tau^{-}$ & $1.05$, $0.95$ \\
                \qquad optimizer & IPOPT \\ \hline
\end{tabular} 
\end{table}

\begin{figure}
\centerline{\includegraphics[scale = 0.45]{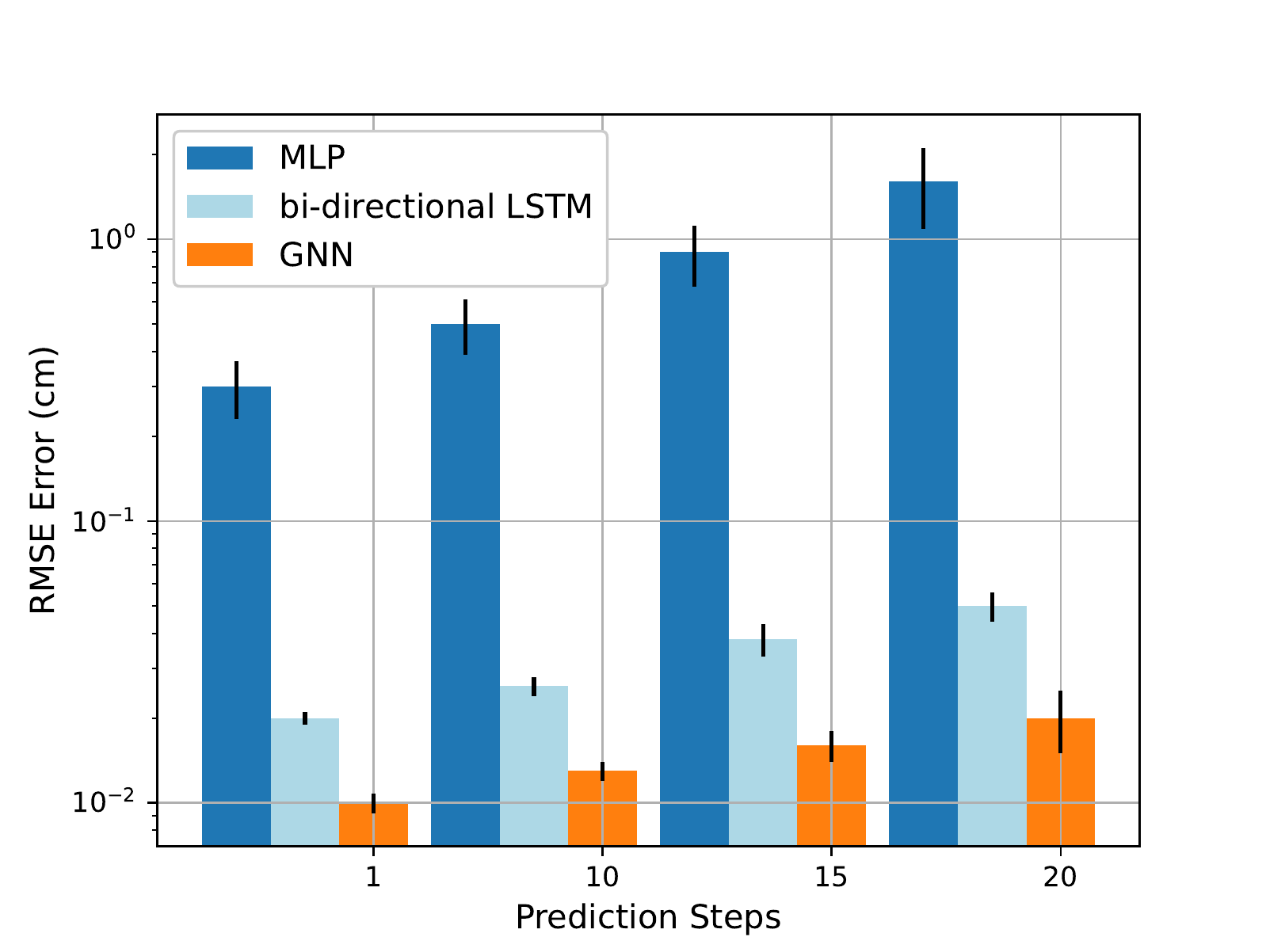}}
\caption{The relation between prediction error and rollout steps of various offline model. The models are forwarded 1 step to 20 steps with the same cable that utilized for training. The y axis is log scaled.}
\label{fig_offline_model_accuracy}
\end{figure}

\subsection{Simulation and Experiment Setup}
\begin{figure*} [!ht]
\centerline{\includegraphics[scale = 0.38]{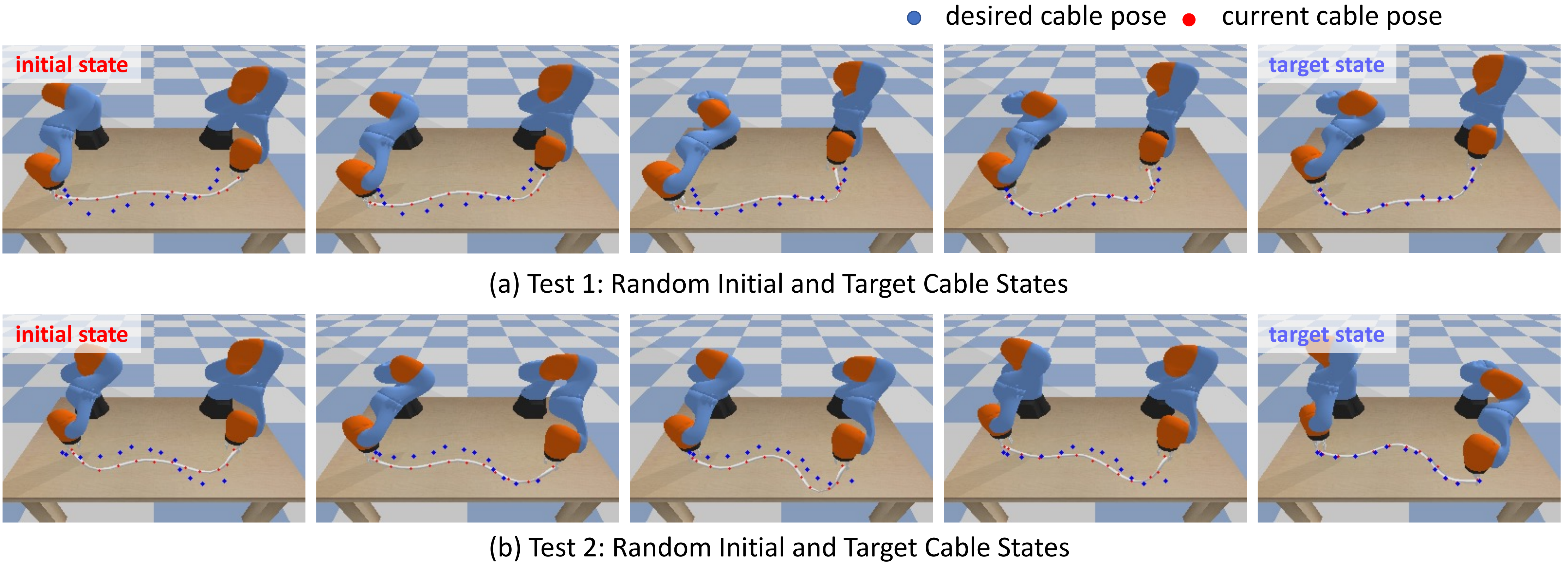}}
\caption{Snapshots of the proposed method. The proposed method is tested to manipulate the cable from random initial shapes to randomly generated targets. With the GNN model, the robot is able to find a rough global trajectory, while the online residual model is able to refine the local behaviors.
}
\label{fig_simulation_sequence}
\end{figure*}

1) \textbf{Simulation Setup:} The simulation setup is shown in Fig.~\ref{fig_simulation_sequence}, two KUKA robots move the ends of a cable so that the cable matches the desired shape. The simulation environment is built upon the PyBullet Physics Engine. The cable model is generated by Gmsh and simulated by the built-in finite element method (FEM). Detailed cable simulation parameters are summarized in Table~\ref{table2}.

\begin{figure}
\centerline{\includegraphics[scale = 0.36]{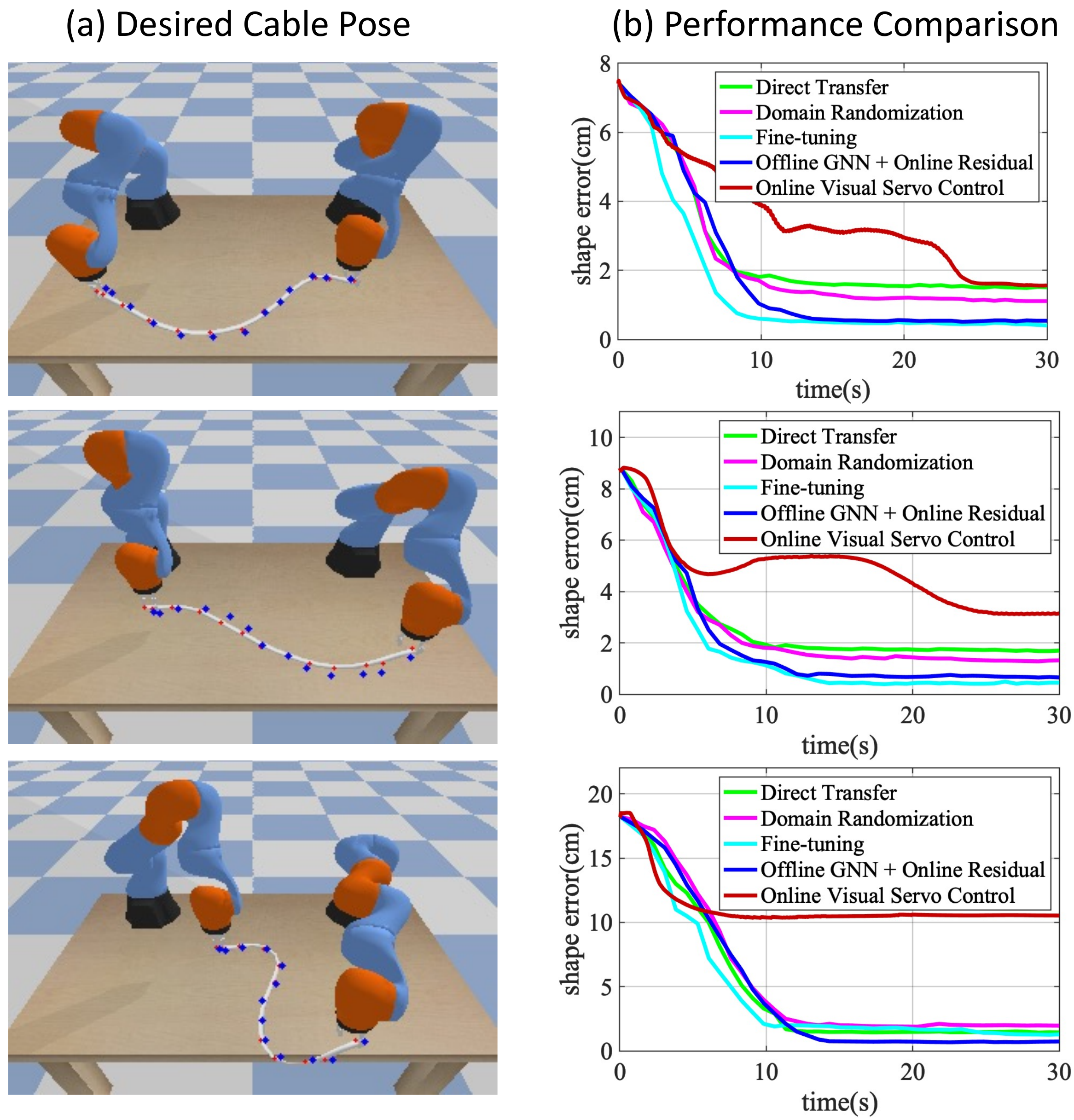}}
\caption{Benchmark comparison results. The robots will manipulate the cable from a straight line to three desired shapes. The curve shows the result in the environment that we double the cable stiffness.
}
\label{fig_simu_benchmark}
\end{figure}

\begin{table*}
    \centering
    \caption{Manipulation Performance Comparison of in Simulation}  
    \label{tab:table3}
   \begin{tabular}{|c|c|c|c|c|c|c|}
  \cline{1-7} 
   \multicolumn{1}{|c|} {\multirow{2}{*}{}} & \multicolumn{2}{c|}{Scenario 1: U shape} &  \multicolumn{2}{c|}{Scenario 2: S shape}  &  \multicolumn{2}{c|}{Scenario 3: Z shape} \\ 
    \cline{2-7}
    \multicolumn{1}{|c|}{}&Terminal Error(cm) & Settling Time(s) & Terminal Error(cm) &Settling Time(s) & Terminal Error(cm) & Settling Time(s)\\
    \cline{1-7}
     \multirow{1}{*}{Online Visual Servo}  & $1.96 \pm 0.29$ & $27.47 \pm 5.44$& $3.28 \pm 0.82$ & $24.76 \pm 3.27$ & $8.72 \pm 1.15$ & $19.43 \pm 6.11$\\
     \cline{1-7}
    \multirow{1}{*}{Direct Transfer GNN}  & $1.54 \pm 0.53$ & $11.32 \pm 1.46$ & $1.61 \pm 0.42$ & $\bm{13.98 \pm 1.31}$ & $1.87 \pm 0.56$ & $11.91 \pm 1.75$\\
    \cline{1-7} 
    \multirow{1}{*}{Domain Randomization}  & $1.11 \pm 0.21$ & $12.60 \pm 1.32$ & $1.30 \pm 0.33$ & $14.33 \pm 1.47$ & $1.76 \pm 0.41$ & $12.07 \pm 1.66$\\
    \cline{1-7} 
    \multirow{1}{*}{GNN + Fine-tune}  & $\bm{0.42 \pm 0.10}$& $\bm{8.31 \pm 0.72}$& $\bm{0.57\pm 0.08}$ & $13.58 \pm 1.06$& $1.19 \pm 0.23$ & $\bm{9.82\pm 1.32}$\\
       \cline{1-7}         
   \multirow{1}{*}{\textbf{GNN + Online Residual}}  & $0.58 \pm 0.13$ & $12.35 \pm 1.09$ & $0.61 \pm 0.14$ & $13.79 \pm 1.50$ & $\bm{0.78 \pm 0.17}$& $16.32 \pm 1.15$\\
    \cline{1-7}
    \end{tabular}
    \footnotesize{* Fine-tuning requires additional data collection before the robot execution, and its performance depends on the quality of the online dataset.}\\
    \end{table*}

\begin{table*}
    \centering
    \caption{Performance Comparison in Experiments with the Ethernet Cable}  
    \label{tab:table4}
   \begin{tabular}{|c|c|c|c|c|}
  \cline{1-5} 
   \multicolumn{1}{|c|} {\multirow{2}{*}{}} & \multicolumn{2}{c|}{Scenario 1: U shape} &  \multicolumn{2}{c|}{Scenario 2: S shape} \\
    \cline{2-5}
    \multicolumn{1}{|c|}{}&Terminal Error(cm) & Settling Time(s) & Terminal Error(cm) &Settling Time(s)\\
    \cline{1-5}
     \multirow{1}{*}{Online Visual Servo}  & $5.58 \pm 1.11$ & $52.04 \pm 12.11$& $1.64 \pm 0.28$ & $44.18 \pm 8.00$\\
    \cline{1-5} 
    \multirow{1}{*}{Direct Transfer GNN}  & $2.54 \pm 0.61$& $25.23 \pm 5.19$& $1.57 \pm 0.35$ & $24.63 \pm 5.16$\\
   \cline{1-5} 
    \multirow{1}{*}{GNN + Domain Randomization}  & $1.91 \pm 0.59$ & $36.29 \pm 5.45$ & $1.36 \pm 0.34$ & $\bm{16.73 \pm 4.37}$ \\
    \cline{1-5} 
     \multirow{1}{*}{GNN + Fine-tune}  & $1.56 \pm 0.43 $ & $\bm{22.29 \pm 5.63}$& $1.12 \pm 0.33$ & $23.03 \pm 4.80$ \\
    \cline{1-5} 
     \multirow{1}{*}{\textbf{GNN + Online Residual}}  & $\bm{1.03 \pm 0.24}$ & $24.29 \pm 7.92$& $\bm{0.99 \pm 0.16}$ & $19.29 \pm 7.71$ \\
    \cline{1-5} 
    \end{tabular}
    \end{table*}
    
\begin{table*}
    \centering
    \caption{Performance Comparison in Experiments with the USB Cable}  
    \label{tab:table5}
   \begin{tabular}{|c|c|c|c|c|}
  \cline{1-5} 
   \multicolumn{1}{|c|} {\multirow{2}{*}{}} & \multicolumn{2}{c|}{Scenario 1: U shape} &  \multicolumn{2}{c|}{Scenario 2: S shape} \\
    \cline{2-5}
    \multicolumn{1}{|c|}{}&Terminal Error(cm) & Settling Time(s) & Terminal Error(cm) &Settling Time(s)\\
    \cline{1-5} 
    \multirow{1}{*}{Online Visual Servo}  & $3.41 \pm 0.91$ & $113.66 \pm 20.93$ & $2.04 \pm 0.16$ & $106.12 \pm 17.09$\\
    \cline{1-5} 
    \multirow{1}{*}{Direct Transfer GNN}  & $1.79 \pm 0.45$& $21.16 \pm 4.38$ & $1.91 \pm 0.67$ & $23.71 \pm 5.13$\\
   \cline{1-5}         
   \multirow{1}{*}{GNN + Domain Randomization}  & $1.39 \pm 0.40$ & $\bm{13.65 \pm 4.39}$ & $1.33 \pm 0.43$ & $31.19 \pm 6.90$ \\
   \cline{1-5}         
   \multirow{1}{*}{GNN + Fine-tune}  & $0.86 \pm 0.38$ & $14.83 \pm 3.87$ & $1.08 \pm 0.24$ & $\bm{21.44 \pm 4.06}$ \\
   \cline{1-5}         
   \multirow{1}{*}{\textbf{GNN + Online Residual}}  & $\bm{0.93 \pm 0.23}$ & $18.81 \pm 7.64$ & $\bm{0.82 \pm 0.16}$ & $34.81 \pm 8.61$ \\
      \cline{1-5}   
    \end{tabular}
    \end{table*}

2) \textbf{Experiment Setup:} The experiment setup is shown in Fig.~\ref{fig_framework}. Two FANUC LR-Mate 200iD robots collaboratively manipulate the cable to different desired shapes. An Intel Realsense RGB-D camera is utilized to obtain the cable point clouds. The online model learning and MPC calculation are achieved on a Ubuntu 18.04 desktop. The optimized robot velocities are sent to the robot for execution with a communication frequency of $100Hz$.

\subsection{Data Collection and Learning}

\textbf{Offline Data Collection and Training:}
The state of the cable is approximated by a series of key points. In simulation, we uniformly select $13$ points on the cable. The cable is initialized with a straight line, and we randomly move the robot end-effector to obtain a trajectory $\{X(t),R(t)\}_{t=0,1,\cdots,200}$, where the trajectory contains $200$ steps of transitions. With this mechanism, we form a data-set with $10K$ trajectories. The network structure and parameters are summarized in Table~\ref{table2}.

\textbf{Online Data Collection and Learning:}
In the online phase, the proposed approach does not require to pre-collect data before execution. The online linear residual model is obtained in real-time by solving (\ref{eqn:least_square_solution}) with $5$ previous cable states and robot movements.

\subsection{Simulation Results}

We first evaluate the performance of the offline model. The cable is randomly initialized, and two robots will randomly move two cable tips for $1$ to $20$ steps. We tested the GNN model with two baselines: 1) a two-layer MLP with the hidden layer size of $128$, 2) a bi-directional LSTM model proposed in~\cite{yan2020self}. All methods are trained with the same amount of data. The results are provided in Fig.~\ref{fig_offline_model_accuracy}. The GNN and bi-directional LSTM models outperform the MLP by a large margin (over $10$ times). The result confirms our hypothesis that the structural designs of the GNN and bi-directional LSTM encode the prior knowledge on how the interaction will transmit inside the cable. Those prior knowledge can make the training result better. While the LSTM model shares many similarities with the GNN on how the interaction is propagated through the vertex, the GNN model performs better in our scenarios. The average prediction error of GNN is less than $2\times 10^{-2}$cm after $20$ steps of predictions, which confirms our assumption that an offline GNN model is a good approximation of the ground truth cable model.

For manipulation, we change the stiffness of the cable ($0.1\times$, $0.5\times$, $1\times$, $2\times$, and $3\times$ of the original values) to test the performance of the proposed online residual model. We compare the performance of the proposed method with three sim-to-real baselines: 1) direct transfer, 2) domain randomization, and 3) fine-tuning the network. To be more specific, direct transfer refers to applying the offline learned model directly to the testing scenarios without modification. Domain randomization denotes randomly perturbs some parameters of the cable in the training dataset. We change the spring elastic stiffness, damping stiffness, and bending stiffness in the range of $0.1$ to $3$ times the original values. For fine-tuning, we collect $2k$ transitions in the testing scenario and re-train the network before executing the robot.
In addition, we also compared with an existing online visual servo control method proposed in~\cite{zhu2018dual,jin2019robust}, where a linear deformation Jacobian is online estimated via least squares.
The benchmark scenarios are illustrated in Fig.~\ref{fig_simu_benchmark}. The robots need to collaboratively manipulate the cable from straight lines to three desired shapes. The results are averaged over $5$ trails with the cable stiffness changing from $0.1$ to $3$ times the original value.

The results are summarized in Table~\ref{tab:table3}. For settling time, the methods with offline model outperform the online visual servo method. Since the online visual servo method purely relies on the local information, it is easy to get stuck in a local optimal region and spend lots of time to get out. This phenomenon is also revealed in Fig.~\ref{fig_simu_benchmark}. In all three scenarios, the online visual servo controller is able to reduce the shape error quickly in the beginning but fails to continue this trend in the end. However, since the visual servo control does not require offline training, the online visual servo method still has the inherent advantage on data efficiency.

For the terminal shape error, we notice that both domain randomization and fine-tuning can improve the test time performance than direct transfer. It is reasonable since the training data for both approaches covers more similar scenarios as in the test time, especially for fine-tuning, which utilizes the data collected in the test scenario to refine the network. However, it is worth noticing that the proposed online residual learning method does not require additional data, and it is robust across all the scenarios. Though fine-tuning slightly outperforms the proposed approach in the first two scenarios, it has a larger error in the third. This may indicate the performance of fine-tuning may greatly depend on the fine-tune dataset.

We also applied the proposed framework in more challenging scenarios as shown in Fig.~\ref{fig_simulation_sequence}, where the initial state and the target state are randomly generated in simulation. Both tasks require large deformation and accurate local adjustment. Utilizing the proposed method, we are able to complete the tasks robustly.

\subsection{Experiment Results}
\begin{figure}
\centerline{\includegraphics[scale = 0.23]{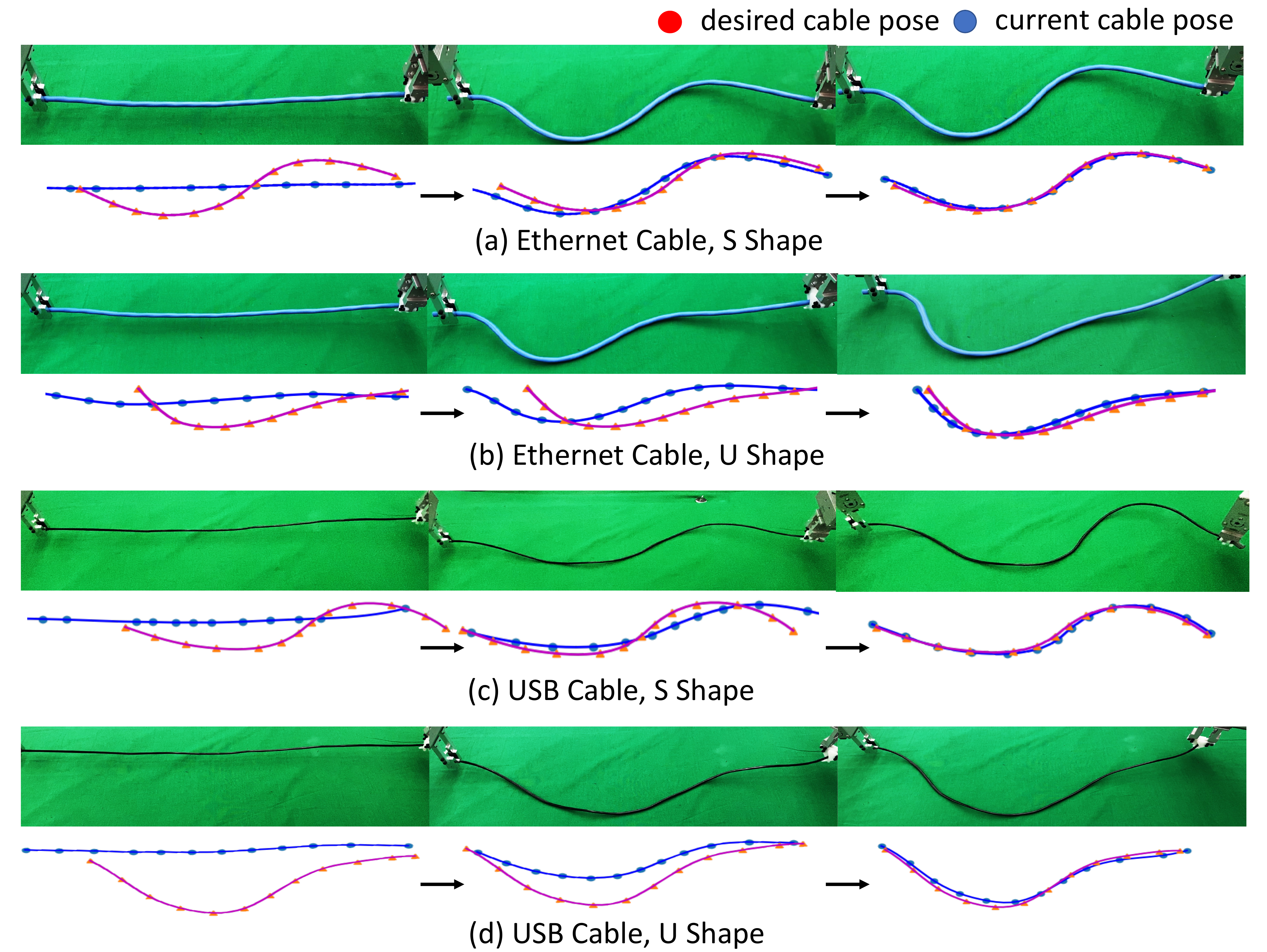}}
\caption{Snapshots of the proposed methods. We tested the performance of the proposed method on two cables. For each cable, we set two desired shapes: U shape and S shape. The proposed method is able to achieve high accuracy efficiently.}
\label{fig_experiment_sequence}
\end{figure}

\begin{figure}
\centerline{\includegraphics[scale = 0.28]{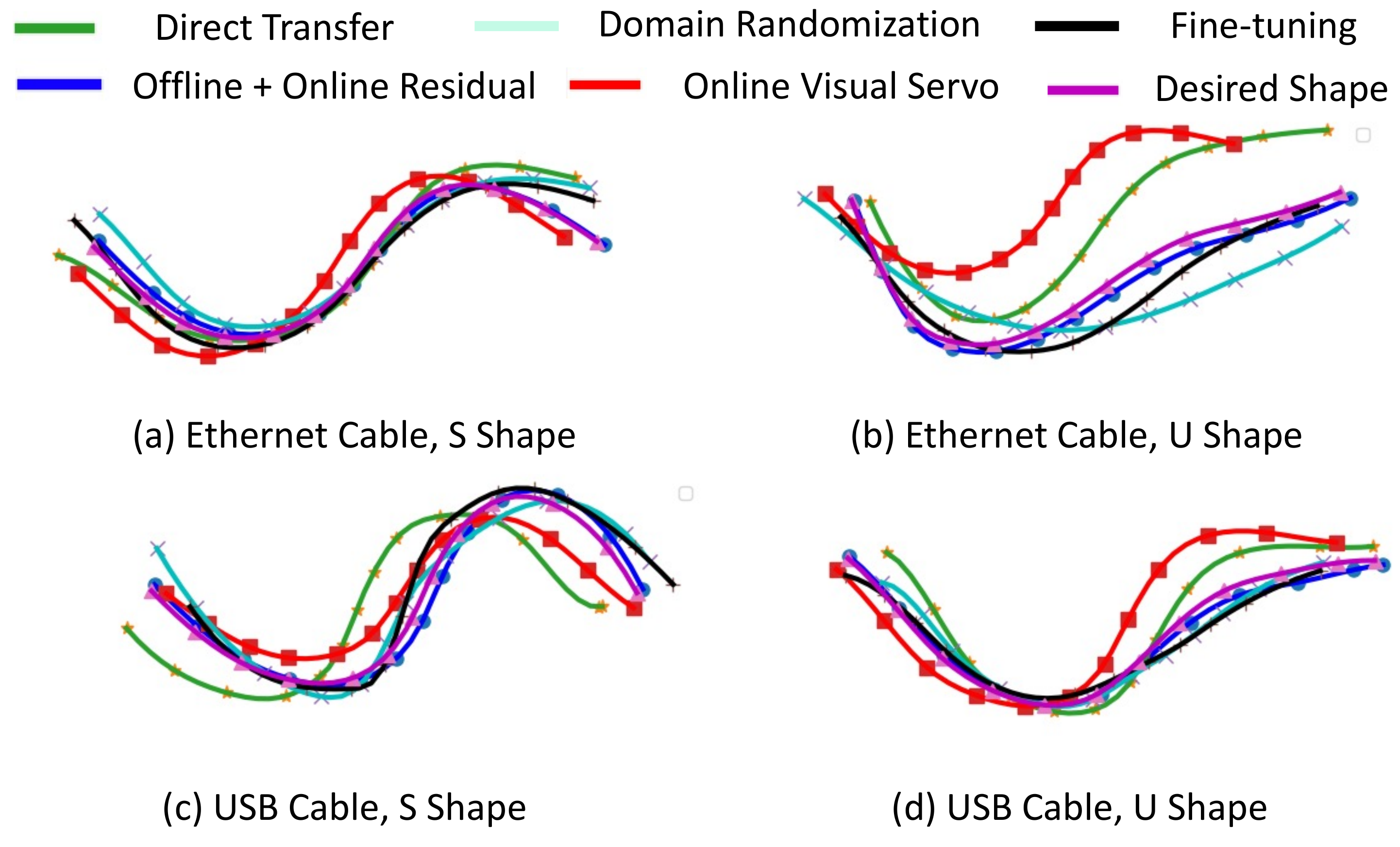}}
\caption{We show the final shapes that each method achieved for the benchmark. The benchmark is to manipulate the cable to a U shape and an S shape, as shown with the purple curve. The proposed method can achieve high accuracy across all the four tasks.
}
\label{fig_final_shape}
\end{figure}

We further evaluate the performance of the proposed framework on two different cables, a blue Ethernet cable, and a black USB cable. The offline GNN dynamics are the same as the offline dynamics in the previous experiment. The online models are initialized by randomly moving around the initial configuration, and the fine-tune dataset is similarly collected with $2k$ transitions. The desired shapes are selected as U shapes and S shapes as the purple curves in Fig.~\ref{fig_final_shape}.

Table~\ref{tab:table4} and Table~\ref{tab:table5} summarize the performance on both cables. The online visual servo control method achieves good performance by gradually learning the local model and moving the cable. However, the online method may converge to a local optimal region and get stuck. As shown in Fig.~\ref{fig_final_shape}(b), the online method fails to achieve the U shape and gets stuck in an S shape that is relative far away from the desire. The offline GNN method has fair performances on both cables. The global model can efficiently guide the robot to avoid potential local optimal regions for better performance. However, due to the sim-to-real gap, the predictions from the offline model may become inaccurate, and the final shape errors are inevitable. Domain randomization can slightly perform better because the training data already includes the deformation of some higher stiffness cables. However, there is still a gap between simulation and reality. Fine-tuning performs better than domain randomization but is not comparable with the proposed approach. As we analyzed in the previous simulation benchmark, its performance may greatly depend on the fine-tune dataset. Since it is expensive and time-consuming to obtain the real-world data, it is challenging to obtain a dataset covering most of the scenarios.
The proposed method achieves the best performance in terms of the terminal error. The result corroborates that the proposed framework is effective and robust in manipulating cables.

\section{Discussion and Future Works}

We proposed a dynamics learning framework for precise cable manipulation, and we demonstrated the effectiveness of the proposed approach mainly in 2D and uncluttered environments. In this section, we discuss the applicability of our approach to more challenging scenarios and point out future research directions.

1) Multi-step cable manipulation: Real-world manipulation tasks may require multi-steps of cable shaping, where the robots need to change the grasping point during the manipulation. 
Finding the optimal grasping point for deformable objects is a challenging research problem, and it is widely studied in the literature~\cite{seita2021learning,yan2020learning}. We plan to utilize the Transporter Network~\cite{seita2021learning} to learn the grasping point from images and combine it with our approach for manipulation.

2) 3D scenarios: The main challenge to transfer the proposed framework from 2D to 3D is the prediction accuracy of the offline model. It is challenging for the offline GNN model to precisely predict the twisting behaviors, as shown in a recent paper~\cite{yang2021learning}. Therefore, we plan to incorporate their modification on the offline model structure and utilize the proposed residual learning approach to manipulate the cable in 3D.

3) Cluttered environment: To the author's knowledge, few model-based approaches have been proposed in the literature due to the complexity of this problem. Mitrano et al.~\cite{mitrano2021learning} proposed to train a classifier to determine where the offline learned model is in effect and utilize a learned policy to recover the robot when the model is not reliable. However, this approach cannot deal with the scenario that the cable requires to contact with the environment. We hypothesize that our proposed online residual learning method can be a potential way to tackle this scenario. The residual model can serve as a contact model, and we leave it as a future research direction.

\bibliographystyle{IEEEtran}
\bibliography{Reference}	

\begin{thebibliography}{10}
\providecommand{\url}[1]{#1}
\csname url@samestyle\endcsname
\providecommand{\newblock}{\relax}
\providecommand{\bibinfo}[2]{#2}
\providecommand{\BIBentrySTDinterwordspacing}{\spaceskip=0pt\relax}
\providecommand{\BIBentryALTinterwordstretchfactor}{4}
\providecommand{\BIBentryALTinterwordspacing}{\spaceskip=\fontdimen2\font plus
\BIBentryALTinterwordstretchfactor\fontdimen3\font minus
  \fontdimen4\font\relax}
\providecommand{\BIBforeignlanguage}[2]{{%
\expandafter\ifx\csname l@#1\endcsname\relax
\typeout{** WARNING: IEEEtran.bst: No hyphenation pattern has been}%
\typeout{** loaded for the language `#1'. Using the pattern for}%
\typeout{** the default language instead.}%
\else
\language=\csname l@#1\endcsname
\fi
#2}}
\providecommand{\BIBdecl}{\relax}
\BIBdecl

\bibitem{brenner2008mathematical}
S.~C. Brenner, L.~R. Scott, and L.~R. Scott, \emph{The mathematical theory of
  finite element methods}.\hskip 1em plus 0.5em minus 0.4em\relax Springer,
  2008, vol.~3.

\bibitem{yu2021adaptive}
M.~Yu, H.~Zhong, F.~Zhong, and X.~Li, ``Adaptive control for robotic
  manipulation of deformable linear objects with offline and online learning of
  unknown models,'' \emph{arXiv preprint arXiv:2107.00194}, 2021.

\bibitem{hu20193}
Z.~Hu, T.~Han, P.~Sun, J.~Pan, and D.~Manocha, ``3-d deformable object
  manipulation using deep neural networks,'' \emph{IEEE Robotics and Automation
  Letters}, vol.~4, no.~4, pp. 4255--4261, 2019.

\bibitem{yan2020learning}
W.~Yan, A.~Vangipuram, P.~Abbeel, and L.~Pinto, ``Learning predictive
  representations for deformable objects using contrastive estimation,''
  \emph{arXiv preprint arXiv:2003.05436}, 2020.

\bibitem{sanchez2020learning}
A.~Sanchez-Gonzalez, J.~Godwin, T.~Pfaff, R.~Ying, J.~Leskovec, and
  P.~Battaglia, ``Learning to simulate complex physics with graph networks,''
  in \emph{International Conference on Machine Learning}.\hskip 1em plus 0.5em
  minus 0.4em\relax PMLR, 2020, pp. 8459--8468.

\bibitem{nair2017combining}
A.~Nair, D.~Chen, P.~Agrawal, P.~Isola, P.~Abbeel, J.~Malik, and S.~Levine,
  ``Combining self-supervised learning and imitation for vision-based rope
  manipulation,'' in \emph{2017 IEEE international conference on robotics and
  automation (ICRA)}.\hskip 1em plus 0.5em minus 0.4em\relax IEEE, 2017, pp.
  2146--2153.

\bibitem{yan2020self}
M.~Yan, Y.~Zhu, N.~Jin, and J.~Bohg, ``Self-supervised learning of state
  estimation for manipulating deformable linear objects,'' \emph{IEEE robotics
  and automation letters}, vol.~5, no.~2, pp. 2372--2379, 2020.

\bibitem{li2018learning}
Y.~Li, J.~Wu, R.~Tedrake, J.~B. Tenenbaum, and A.~Torralba, ``Learning particle
  dynamics for manipulating rigid bodies, deformable objects, and fluids,''
  \emph{arXiv preprint arXiv:1810.01566}, 2018.

\bibitem{pfaff2020learning}
T.~Pfaff, M.~Fortunato, A.~Sanchez-Gonzalez, and P.~W. Battaglia, ``Learning
  mesh-based simulation with graph networks,'' \emph{arXiv preprint
  arXiv:2010.03409}, 2020.

\bibitem{tobin2017domain}
J.~Tobin, R.~Fong, A.~Ray, J.~Schneider, W.~Zaremba, and P.~Abbeel, ``Domain
  randomization for transferring deep neural networks from simulation to the
  real world,'' in \emph{2017 IEEE/RSJ international conference on intelligent
  robots and systems (IROS)}.\hskip 1em plus 0.5em minus 0.4em\relax IEEE,
  2017, pp. 23--30.

\bibitem{chebotar2019closing}
Y.~Chebotar, A.~Handa, V.~Makoviychuk, M.~Macklin, J.~Issac, N.~Ratliff, and
  D.~Fox, ``Closing the sim-to-real loop: Adapting simulation randomization
  with real world experience,'' in \emph{2019 International Conference on
  Robotics and Automation (ICRA)}.\hskip 1em plus 0.5em minus 0.4em\relax IEEE,
  2019, pp. 8973--8979.

\bibitem{ramos2019bayessim}
F.~Ramos, R.~C. Possas, and D.~Fox, ``Bayessim: adaptive domain randomization
  via probabilistic inference for robotics simulators,'' \emph{arXiv preprint
  arXiv:1906.01728}, 2019.

\bibitem{ajay2018augmenting}
A.~Ajay, J.~Wu, N.~Fazeli, M.~Bauza, L.~P. Kaelbling, J.~B. Tenenbaum, and
  A.~Rodriguez, ``Augmenting physical simulators with stochastic neural
  networks: Case study of planar pushing and bouncing,'' in \emph{2018 IEEE/RSJ
  International Conference on Intelligent Robots and Systems (IROS)}.\hskip 1em
  plus 0.5em minus 0.4em\relax IEEE, 2018, pp. 3066--3073.

\bibitem{kloss2017combining}
A.~Kloss, S.~Schaal, and J.~Bohg, ``Combining learned and analytical models for
  predicting action effects,'' \emph{arXiv preprint arXiv:1710.04102}, vol.~11,
  2017.

\bibitem{navarro2013visually}
D.~Navarro-Alarcon, Y.~Liu, J.~G. Romero, and P.~Li, ``Visually servoed
  deformation control by robot manipulators,'' in \emph{2013 IEEE International
  Conference on Robotics and Automation}.\hskip 1em plus 0.5em minus
  0.4em\relax IEEE, 2013, pp. 5259--5264.

\bibitem{zhu2018dual}
J.~Zhu, B.~Navarro, P.~Fraisse, A.~Crosnier, and A.~Cherubini, ``Dual-arm
  robotic manipulation of flexible cables,'' in \emph{2018 IEEE/RSJ
  International Conference on Intelligent Robots and Systems (IROS)}.\hskip 1em
  plus 0.5em minus 0.4em\relax IEEE, 2018, pp. 479--484.

\bibitem{zhu2021vision}
J.~Zhu, D.~Navarro-Alarcon, R.~Passama, and A.~Cherubini, ``Vision-based
  manipulation of deformable and rigid objects using subspace projections of 2d
  contours,'' \emph{Robotics and Autonomous Systems}, vol. 142, p. 103798,
  2021.

\bibitem{jin2019robust}
S.~Jin, C.~Wang, and M.~Tomizuka, ``Robust deformation model approximation for
  robotic cable manipulation,'' in \emph{2019 IEEE/RSJ International Conference
  on Intelligent Robots and Systems (IROS)}.\hskip 1em plus 0.5em minus
  0.4em\relax IEEE, 2019, pp. 6586--6593.

\bibitem{tang2018track}
T.~Tang and M.~Tomizuka, ``Track deformable objects from point clouds with
  structure preserved registration,'' \emph{The International Journal of
  Robotics Research}, p. 0278364919841431, 2018.

\bibitem{tang2018framework}
T.~Tang, C.~Wang, and M.~Tomizuka, ``A framework for manipulating deformable
  linear objects by coherent point drift,'' \emph{IEEE Robotics and Automation
  Letters}, vol.~3, no.~4, pp. 3426--3433, 2018.

\bibitem{wang2021trajectory}
C.~Wang, J.~Bingham, and M.~Tomizuka, ``Trajectory splitting: A distributed
  formulation for collision avoiding trajectory optimization,'' in \emph{2021
  IEEE/RSJ International Conference on Intelligent Robots and Systems
  (IROS)}.\hskip 1em plus 0.5em minus 0.4em\relax IEEE, 2021, pp. 8113--8120.

\bibitem{wachter2006implementation}
A.~W{\"a}chter and L.~T. Biegler, ``On the implementation of an interior-point
  filter line-search algorithm for large-scale nonlinear programming,''
  \emph{Mathematical programming}, vol. 106, no.~1, pp. 25--57, 2006.

\bibitem{wachter2005line}
A.~Wachter and L.~T. Biegler, ``Line search filter methods for nonlinear
  programming: Motivation and global convergence,'' \emph{SIAM Journal on
  Optimization}, vol.~16, no.~1, pp. 1--31, 2005.

\bibitem{seita2021learning}
D.~Seita, P.~Florence, J.~Tompson, E.~Coumans, V.~Sindhwani, K.~Goldberg, and
  A.~Zeng, ``Learning to rearrange deformable cables, fabrics, and bags with
  goal-conditioned transporter networks,'' in \emph{2021 IEEE International
  Conference on Robotics and Automation (ICRA)}.\hskip 1em plus 0.5em minus
  0.4em\relax IEEE, 2021, pp. 4568--4575.

\bibitem{yang2021learning}
Y.~Yang, J.~A. Stork, and T.~Stoyanov, ``Learning to propagate interaction
  effects for modeling deformable linear objects dynamics,'' in \emph{2021 IEEE
  International Conference on Robotics and Automation (ICRA)}.\hskip 1em plus
  0.5em minus 0.4em\relax IEEE, 2021, pp. 1950--1957.

\bibitem{mitrano2021learning}
P.~Mitrano, D.~McConachie, and D.~Berenson, ``Learning where to trust
  unreliable models in an unstructured world for deformable object
  manipulation,'' \emph{Science Robotics}, vol.~6, no.~54, p. eabd8170, 2021.

\end{thebibliography}
\vspace{12pt}
\end{document}